\documentclass{article} 
\usepackage{iclr2027_conference,times}

\usepackage{amsmath,amsfonts,bm}

\def\eqref#1{equation~\ref{#1}}

\def\1{\bm{1}}

\DeclareMathAlphabet{\mathsfit}{\encodingdefault}{\sfdefault}{m}{sl}
\SetMathAlphabet{\mathsfit}{bold}{\encodingdefault}{\sfdefault}{bx}{n}

\usepackage{hyperref}
\usepackage{url}
\usepackage{xcolor}
\usepackage{colortbl}
\usepackage{booktabs}
\usepackage{graphicx}
\usepackage{needspace}
\usepackage{float}
\usepackage{capt-of}
\usepackage{caption}
\usepackage[normalem]{ulem}

\title{Ada-BFM: Adapting Behavior Foundation Models for Humanoid Scene Interaction}

\author{%
\begin{minipage}{\dimexpr\textwidth-2\tabcolsep\relax}
\centering
\small
\bfseries
Furui Xu$^{1,2*}$ \quad
Weishuai Zeng$^{2,3*}$ \quad
Zirui Wang$^{2,4*}$ \quad
Kangning Yin$^{2,5}$ \\
Jiacheng Bao$^{2}$ \quad
Huayi Wang$^{2}$ \quad
Yingying Qin$^{6}$ \quad
Yufei Xue$^{2}$ \quad
Yang Tang$^{2}$ \\
Zenghao Tang$^{2}$ \quad
Kailin Li$^{2\dagger}$ \quad
Jingbo Wang$^{2\dagger}$ \quad
Weinan Zhang$^{2,5\dagger}$ \\
\normalfont
\footnotesize
$^{1}$Fudan University \hspace{0.1em}
$^{2}$Shanghai Artificial Intelligence Laboratory \hspace{0.1em}
$^{3}$The Chinese University of Hong Kong \\
$^{4}$Zhejiang University \hspace{0.1em}
$^{5}$Shanghai Jiao Tong University \hspace{0.1em}
$^{6}$Southeast University \\
$^{*}$Equal contribution. \quad $^{\dagger}$Corresponding authors.
\end{minipage}%
}

\newcommand{\succdrop}[1]{$_{\downarrow #1}$}

\iclrfinalcopy 
\begin{document}

\maketitle
\vspace*{-12pt}   
\lhead{} 
\vspace*{-8pt}   
\lhead{} 

\begin{figure}[!h]
\begin{center}
\includegraphics[width=\linewidth]{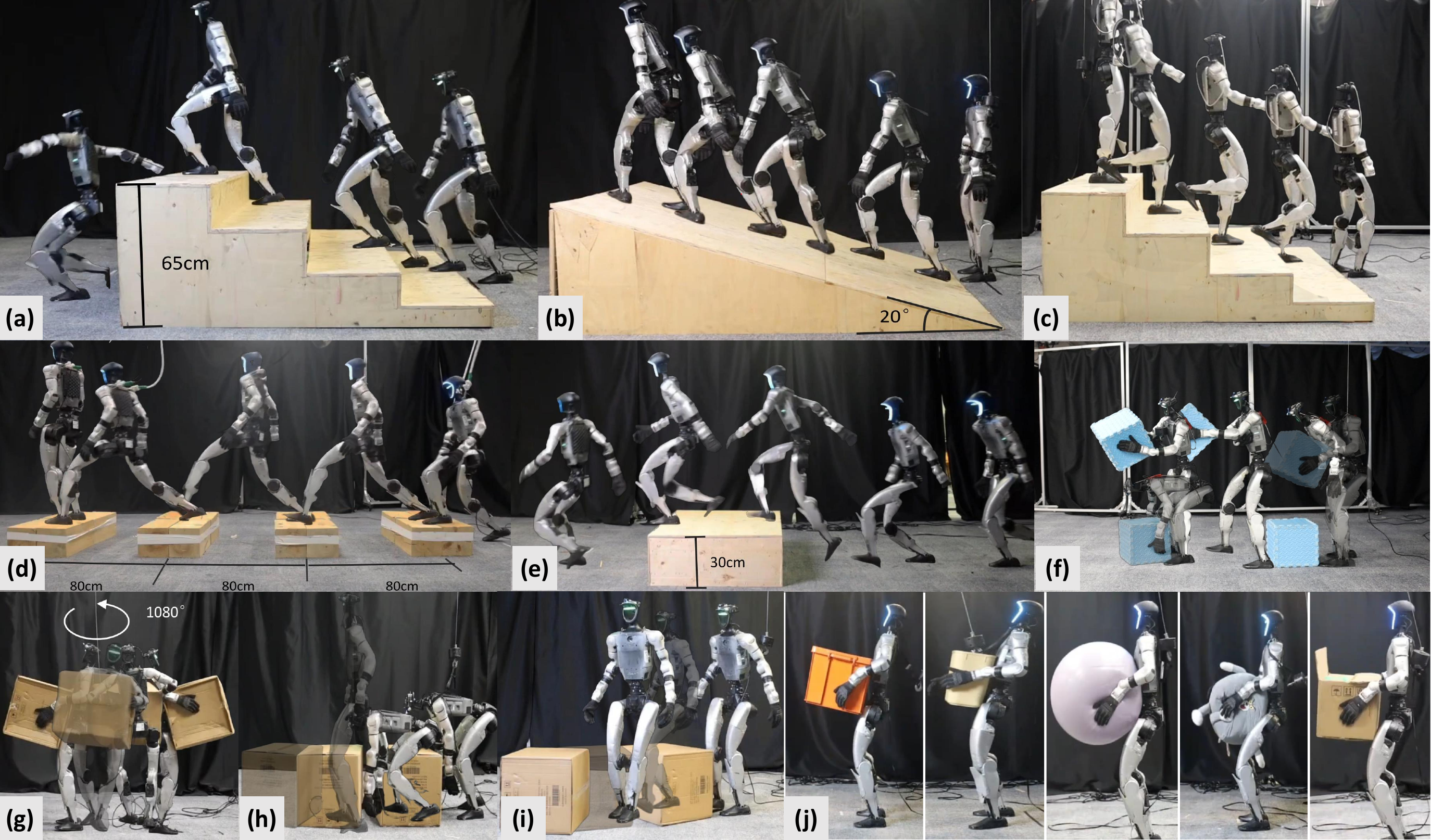}
\end{center}
\caption{\small{\textbf{Ada-BFM} extends the capabilities of pretrained BFMs to humanoid scene interaction, with a single adapted model shared across diverse tasks within each interaction domain. Ada-BFM supports diverse interaction behaviors: (a) stepping up stairs and jumping down; (b) ascending a slope; (c) descending stairs; (d) repeatedly stepping over curbs; (e) stepping over an obstacle; (f) walking while lifting a box; (g) rapidly turning $1080^\circ$ while holding a box; (h) pushing a box; (i) kicking a box; and (j) lifting objects of varying shapes, materials, and weights. Website: \url{https://ada-bfm.github.io/}}}
\label{fig:teaser}
\end{figure}

\begin{abstract}
Behavior Foundation Models (BFMs) have emerged as a promising paradigm for general-purpose humanoid control through pretraining on large-scale behavioral data. However, their proficiency in whole-body coordination does not readily extend to scene interaction, which requires adapting motion to scene-dependent geometric constraints and contact dynamics. To bridge this gap, we introduce \textbf{Ada-BFM}, a post-training framework that adapts pretrained BFMs for humanoid scene interaction. {Ada-BFM} augments the original architecture with a lightweight adapter, enabling interaction-specific adjustments while leveraging pretrained control capabilities. Rather than producing a separate controller for each task, the framework yields a shared model for each interaction domain, supporting diverse tasks involving terrain or object interaction. Extensive experiments in simulation and the real world demonstrate that {Ada-BFM} enables diverse scene interaction behaviors, accelerates learning compared with training from scratch, and largely preserves pretrained behavioral competence. These findings establish {Ada-BFM} as an effective and efficient approach to extending the capabilities of BFMs beyond their pretraining settings while retaining their strengths in general whole-body control.
\end{abstract}

\section{Introduction}
\label{sec:introduction}

Humanoid whole-body control has advanced substantially in recent years, achieving impressive capabilities across diverse motor skills~\citep{ben2025homie,huang2025learning,he2025asap}. However, existing approaches often rely on extensive reward engineering tailored to specific tasks, limiting their versatility and hindering generalization across scenarios. To address these limitations, recent research has adopted behavior learning as a shared objective across whole-body control tasks, giving rise to {Behavior Foundation Models (BFMs)}, large-scale pretrained humanoid controllers that enable robust and versatile control in response to diverse behavioral specifications~\citep{zeng2025behavior,li2026bfm}. Motion tracking~\citep{peng2018deepmimic,chen2025gmt,yin2026unitracker,liao2026beyondmimic} is identified as a unified and scalable task for pretraining these models, allowing BFMs to acquire increasingly rich behavioral knowledge and gain consistent improvements in control capabilities as pretraining scales~\citep{luo2026sonic,zeng2026scaling,chen2026holomotion}.

Despite this growth in behavioral knowledge, existing BFM pretraining remains largely \emph{scene-agnostic}, focusing on the realization of behavioral specifications without explicitly accounting for the surrounding environments. This limits BFMs' ability to interact with terrain and objects, where successful execution requires not only following control commands but adapting motion to scene-dependent geometric constraints and contact dynamics~\citep{omniretarget2025,php2026,hdmi2025,omniconnect2026}. Although a broad repertoire of behaviors does not, by itself, ensure effective scene interaction, the underlying behavioral knowledge provides a potentially valuable foundation for efficient transfer. This motivates scene-aware adaptation of BFMs that builds upon pretrained behavioral capabilities to meet the demands of physical interaction.

To this end, we propose \textbf{Ada-BFM}, a post-training framework that adapts the pretrained BFM for humanoid scene interaction. The framework is designed around three core adaptation objectives: 1) \textit{Effectiveness}, enabling a single adapted model within each interaction domain to handle diverse tasks and outperform baseline approaches; 2) \textit{Efficiency}, leveraging the behavioral knowledge acquired through pretraining to accelerate the acquisition of scene-interaction capabilities; and 3) \textit{Retention}, preserving the broad behavioral repertoire of the pretrained model while minimizing degradation of its original control capabilities. Together, these objectives extend the capabilities of BFMs beyond their pretraining while preserving their strengths as general-purpose humanoid controllers.

As a concrete instantiation, Ada-BFM is built upon a pretrained BFM from ScaleBFM~\citep{zeng2026scaling}. Its \textit{Humanoid Transformer} backbone employs self-attention to capture temporal dependencies within behavioral trajectories and cross-attention to incorporate behavioral specifications that guide action generation. A learnable query token is appended to the end of the context sequence to aggregate control-relevant information and its output representation is used to predict actions. To enable scene-aware adaptation, Ada-BFM uses a lightweight adapter to incorporate scene-relevant information into Transformer blocks. The adapter comprises a gated cross-attention module and a gated feed-forward layer. This design extends the backbone's conditioning mechanism: motion goals specify the intended behavior, while scene observations describe the environment in which it must be executed. Rather than concatenating scene inputs with the original specifications, the adapter incorporates them through a separate conditioning pathway designed to minimize interference with pretrained control capabilities.

To evaluate the performance of the proposed framework in adapting pretrained BFMs for humanoid scene interaction, we consider two representative domains: terrain interaction and object interaction. Within each domain, we train a single adapted model to handle diverse tasks rather than a separate model for each task. Both domains use the same adaptation framework, with domain-specific interaction environments and rewards. Depth images provide a common representation of scene information across both domains. Throughout post-training, the pretrained backbone remains frozen and continues to process control commands, while the trainable adaptation pathway learns to leverage depth perception to modulate the resulting behavior for scene-aware physical interaction.

Comprehensive experiments in simulation and the real world demonstrate that Ada-BFM outperforms baseline approaches on terrain-interaction and object-interaction tasks, accelerates learning of scene-interaction behaviors compared with training from scratch and largely preserves pretrained control capabilities. These findings demonstrate the effectiveness, efficiency and retention of the proposed framework, establishing it as a promising approach for extending BFMs beyond their pretraining settings while retaining their strengths in general whole-body control.

\section{Related Work}

\subsection{General Humanoid Motion Tracking and Behavior Foundation Models}
Physics-based imitation learns whole-body skills from human
motion~\citep{peng2018deepmimic,peng2022ase}, and motion trackers expose them as an
interface for
teleoperation~\citep{luo2023perpetual,he2024omnih2o,yin2026unitracker,chen2025gmt,liao2026beyondmimic}.
Behavior foundation models apply this approach at larger scale, with competence
improving as pretraining
grows~\citep{zeng2025behavior,luo2026sonic,zeng2026scaling,li2026bfm,chen2026holomotion}.
These models achieve strong whole-body motor capability. However, they are pretrained
largely on flat-ground data and therefore lack the capability to interact with terrain
or objects.


\subsection{Humanoid Scene Interaction}
\label{sec:related-hsi}
Existing scene-interaction controllers acquire interaction skills through task-specific
engineering. For terrain, perceptive policies map
depth or voxel observations to actions for a specific
terrain~\citep{php2026,hiking2026,ben2025gallant}, while reference-conditioned trackers
reconcile the reference with the terrain by modifying motion
priors~\citep{wang2026perceptivebfm}, relaxing tracking
constraints~\citep{li2026pgmt}, or supplying scene-aligned 4D
demonstrations~\citep{brandes2026egohtr}. Object interaction is learned from human
videos~\citep{hdmi2025,wang2026humanx}, adversarial motion
priors~\citep{physhsi2025}, or contact
representations~\citep{omniconnect2026,scenebot2026}. Training such a controller from scratch requires acquiring whole-body coordination and
interaction jointly. In practice, it relies on hand-designed rewards or
teacher-student distillation. A pretrained motion prior, by contrast, allows the controller to learn the interaction
skill more efficiently.

\subsection{Post-Training Adaptation}
\label{sec:related-adapter}
Post-training adaptation extends the capabilities of a pretrained model. In language and
vision models, adapters and low-rank updates add a small number of trainable parameters to a fixed
network~\citep{houlsby2019adapters,rebuffi2017residual,hu2022lora,alayrac2022flamingo}.
In humanoid control, GRAFT and C-GMT adapt the reference or the tracking
objective~\citep{li2026graft,parsuram2026compliantgmt}, ViBe adapts perception
features~\citep{krishna2026vibe}, and ResMimic learns a residual over the action
output~\citep{zhao2025resmimic}. Task Tokens and FIBER keep the backbone frozen and add
a conditioning pathway~\citep{vainshtein2025tasktokens,lin2026fiber}, and TokenHSI
unifies multiple interaction tasks for simulated characters through a tokenized
interface~\citep{pan2025tokenhsi}. 
Post-training adaptation must extend the capabilities of the pretrained
model while preserving as much of its original capability as possible.

\section{Method}
\label{sec:method}

Ada-BFM adapts a pretrained ScaleBFM controller~\citep{zeng2026scaling}
to humanoid scene interaction. It keeps the pretrained actor frozen and adds
a trainable perception pathway with a gated adapter. We train a shared
model for each interaction domain.

\subsection{Problem Formulation}
\label{sec:problem-formulation}

Following ScaleBFM~\citep{zeng2025behavior,zeng2026scaling}, we define humanoid
behavior as a trajectory of proprioceptive states $\bm{s}^{p}_{t}$ and actions
$\bm{a}_{t}$ over $T$ control steps:
\begin{equation}
    \mathcal{B} \equiv
    \left(\bm{s}^{p}_{0},\bm{a}_{0},\bm{s}^{p}_{1},\bm{a}_{1},
    \ldots,\bm{s}^{p}_{T-1},\bm{a}_{T-1},\bm{s}^{p}_{T}\right).
    \label{eq:behavior-definition}
\end{equation}
The pretrained controller implements
$\pi_{\bm{\theta}_{0}}(\bm{a}_{t}\mid\mathcal{H}_{t},\mathcal{G}_{t})$,
where $\mathcal{H}_{t}$ is a finite proprioception--action history and
$\mathcal{G}_{t}$ contains future whole-body targets expressed relative to the
robot's root.

Scene interaction additionally requires execution to account for environmental
geometry and contact dynamics. We model each interaction domain $k$ as an MDP
$\mathcal{M}^{k}=\langle\mathcal{S}^{k},\mathcal{A},\mathcal{T}^{k},
\mathcal{R}^{k},\gamma\rangle$. Each domain has its own states, dynamics, and rewards.
All domains share an action space and discount factor $\gamma$. The actor receives the history $\mathcal{H}_{t}$ and motion goal $\mathcal{G}_{t}$, together with a scene
observation $\bm{o}^{k}_{t}$. Whereas $\mathcal{G}_{t}$ specifies the intended
motion, $\bm{o}^{k}_{t}$ describes the scene in which it must be executed.

We add a scene pathway with parameters $\bm{\phi}_{k}$ to the pretrained
controller. It comprises a depth encoder, a tokenizer, and an adapter, yielding
$\pi_{\bm{\theta}_{0},\bm{\phi}_{k}}(\bm{a}_{t}\mid\mathcal{H}_{t},
\mathcal{G}_{t},\bm{o}^{k}_{t})$. A single parameter set $\bm{\phi}_{k}$ is shared
across tasks within each domain. We fix $\bm{\theta}_{0}$, the pretrained network parameters, and optimize the scene pathway for expected
discounted interaction return:
\begin{equation}
    \max_{\bm{\phi}_{k}}\;
    \mathbb{E}_{\pi_{\bm{\theta}_{0},\bm{\phi}_{k}}}
    \left[\sum_{t=0}^{T-1}\gamma^{t}r^{k}_{t}\right].
    \label{eq:adaptation-objective}
\end{equation}
The expectation is over rollouts of length $T$ in $\mathcal{M}^{k}$ under the
adapted policy, and $r^{k}_{t}$ is the domain reward.

\subsection{Learning Paradigm}
\label{sec:learning-paradigm}

\paragraph{Interaction objectives guided by motion.}
We inherit ScaleBFM's motion interface and reward for tracking integrated
whole-body trajectories in the global frame. Tracking root motion together with
body pose provides dense supervision for coordinated behavior during adaptation.
At control step $t$, let $r_t^{\mathrm{BFM}}$ denote the original BFM reward.
The terrain interaction reward $r_t^{\mathrm{terrain}}$ retains this reward
to guide reference motion execution under new geometric and contact constraints.
The object interaction reward $r_t^{\mathrm{object}}$ additionally includes
a task reward $r_t^{\mathrm{task}}$ that supervises the interaction outcome.
Appendix~\ref{app:rewards} details the reward terms. Thus,
\begin{equation}
    r_t^{\mathrm{terrain}}=r_t^{\mathrm{BFM}},\qquad
    r_t^{\mathrm{object}}=r_t^{\mathrm{BFM}}+r_t^{\mathrm{task}}.
    \label{eq:interaction-rewards}
\end{equation}

\paragraph{Policy optimization.}
We train the scene pathway with proximal policy optimization (PPO). The
minimization objective is
\begin{equation}
    \mathcal{L} = \mathcal{L}^{\mathrm{clip}}
    + c_{v}\,\mathcal{L}^{\mathrm{value}}
    - c_{e}\,\mathbb{H}[\pi].
    \label{eq:policy-optimization}
\end{equation}
Here $\mathcal{L}^{\mathrm{clip}}$ is the negative clipped PPO surrogate,
$\mathcal{L}^{\mathrm{value}}$ is clipped value regression, and
$\mathbb{H}[\pi]$ is an entropy bonus supporting exploration. The coefficients \(c_v\) and \(c_e\) weight the value loss and entropy bonus, respectively.
Tracking rewards supervise body motion, and interaction rewards supervise
object outcomes.

\paragraph{Adaptive sampling.}
We adapt the distribution of episode starting points using failures observed
during training. Each reference motion is divided into temporal intervals.
Intervals with larger moving averages of failure counts are sampled more often,
and a starting frame is sampled uniformly within the selected interval.
A uniform component maintains coverage of the remaining intervals.

\paragraph{Asymmetric observations.}
Following ScaleBFM, we use an asymmetric actor--critic. The actor receives
proprioception--action history, motion goals, and depth observations. A separate
critic also receives privileged simulator information about whole-body states
and the terrain or object. This information supports value estimation and is
not required at deployment.  Depth provides a common scene representation for both domains.
Preprocessing details are provided in Appendix~\ref{app:depth-preprocessing}.

\begin{figure}[t]
\begin{center}
\includegraphics[width=\linewidth]{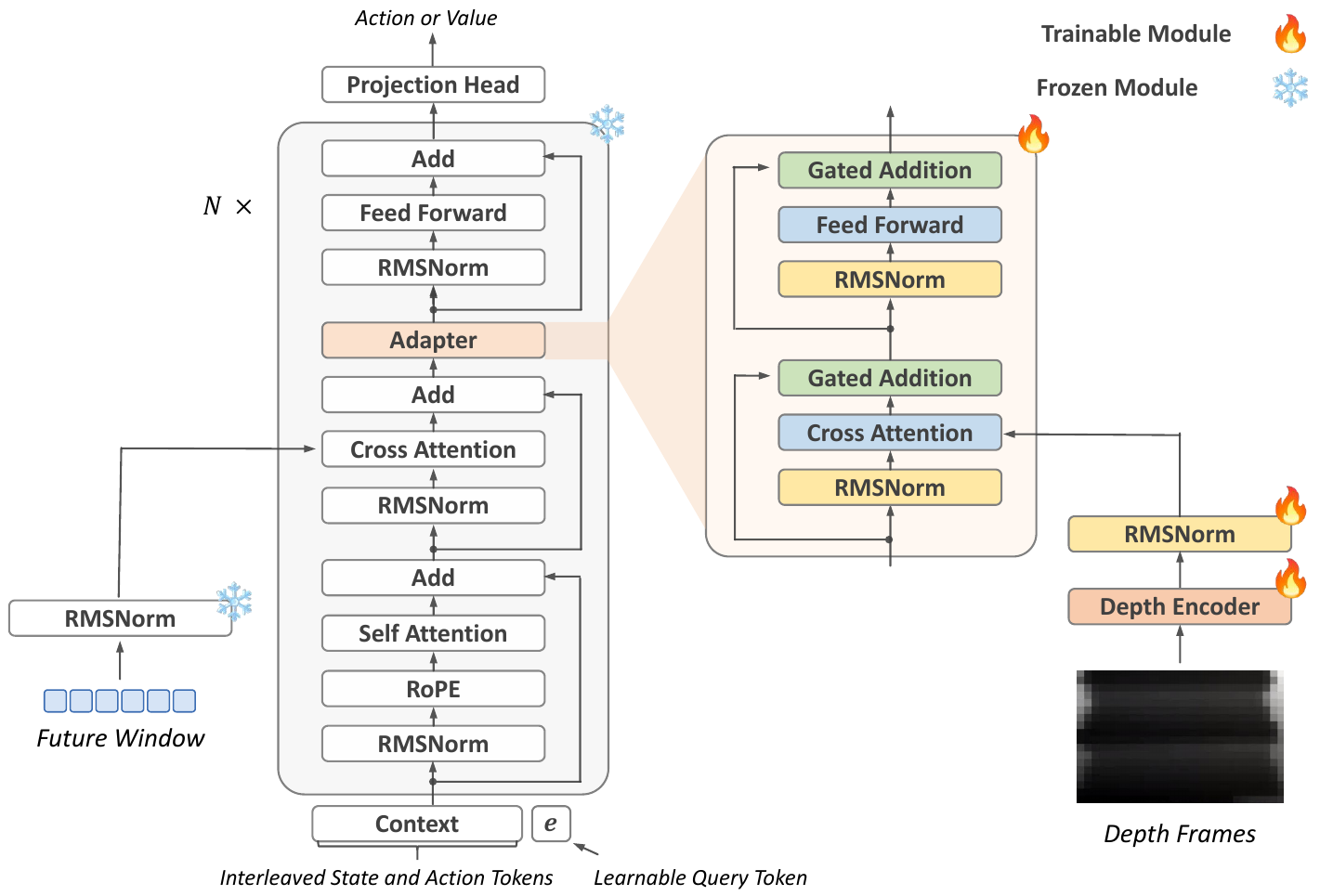}
\end{center}
\caption{Overview of Ada-BFM. Left: we freeze the pretrained BFM backbone. The adapter refines the query token used for action
prediction. Right: a trainable depth encoder supplies scene features to the
adapter. Gated cross-attention adds these features to the query representation,
and a gated feed-forward branch transforms the combined features.}
\label{fig:method-overview}
\end{figure}

\subsection{Model Architecture}
\label{sec:model-architecture}

Ada-BFM augments a frozen BFM with a trainable pathway for scene information.
A depth encoder and tokenizer provide visual features to the adapter,
which refines the action representation while retaining the original motion
conditioning pathway. Figure~\ref{fig:method-overview} illustrates the architecture.

\paragraph{Frozen pretrained backbone.}
Ada-BFM builds on the Humanoid Transformer in ScaleBFM~\citep{zeng2026scaling}.
Throughout adaptation, we freeze all pretrained backbone parameters, including
the tokenizers, the initial query embedding, the Transformer blocks, and the
action head.
The backbone interleaves historical proprioception and action tokens into a
context sequence. A query token initialized with the pretrained embedding
attends to this context and provides the representation used for action
prediction. An attention mask allows the
query token to read the historical context but prevents context tokens from
attending to the query token. This separates the aggregation of information
for action prediction from the processing of historical context.
Reference cross-attention incorporates motion goals encoded by the pretrained
goal tokenizer. The action head maps the final query representation to the
action mean.

\paragraph{Scene representation.}
We use egocentric depth observations to provide a common representation of scene geometry for terrain and object interaction. A trainable
convolutional encoder compresses the depth observation $\bm{o}^{k}_{t}$
into visual features. A linear tokenizer maps these features
to a scene representation $\bm{z}_{t}$. This representation provides perceptual information about the surrounding
terrain or objects, and the adapter incorporates this
information through a separate conditioning pathway.

\paragraph{Gated scene adaptation.}
We insert the adapter after reference cross-attention and before the original
feed-forward network of a Transformer block. In our experiments, we adapt
only the final Transformer block. At this location, the query token
incorporates both historical context and the motion goal. The adapter updates
only this query token, while context tokens follow the original backbone
path. Let $\bm{h}_{t}$ denote the query representation at the adapter input.
The adapter first incorporates scene features through gated cross-attention,
then applies a gated feed-forward transformation:
\begin{equation}
    \begin{aligned}
    \tilde{\bm{h}}_{t} &= \bm{h}_{t}
    + \tanh(\alpha)\,\operatorname{Attn}
    \bigl(N_h(\bm{h}_{t}),N_z(\bm{z}_{t})\bigr),\\
    \bm{h}'_{t} &= \tilde{\bm{h}}_{t}
    + \tanh(\beta)\,\operatorname{FFN}_{\mathrm{adapt}}
    \bigl(N_f(\tilde{\bm{h}}_{t})\bigr).
    \label{eq:adapter-update}
    \end{aligned}
\end{equation}
Here $N_h$, $N_z$, and $N_f$ are separate RMSNorm modules.
The attention query is derived from $\bm{h}_{t}$, while the keys and values
are derived from the scene representation $\bm{z}_{t}$. The adapter has its own
trainable attention projections and normalization parameters.
The attention branch incorporates scene information into the query
representation, and the SwiGLU feed-forward branch further transforms
the resulting representation. Each branch contributes a residual update
scaled by its own learned scalar gate.
The updated query token then passes through the frozen
backbone.

\paragraph{Initialization and trainable components.}
The gates $\alpha$ and $\beta$ are learned scalars initialized to zero.
Both residual branches therefore contribute zero at initialization, so adaptation
starts from the pretrained action mean. We train the depth encoder, scene
tokenizer, and adapter. The original tokenizers, initial query embedding,
Transformer blocks, and action head remain frozen.

\begin{figure}[t]
\begin{center}
\includegraphics[width=\linewidth]{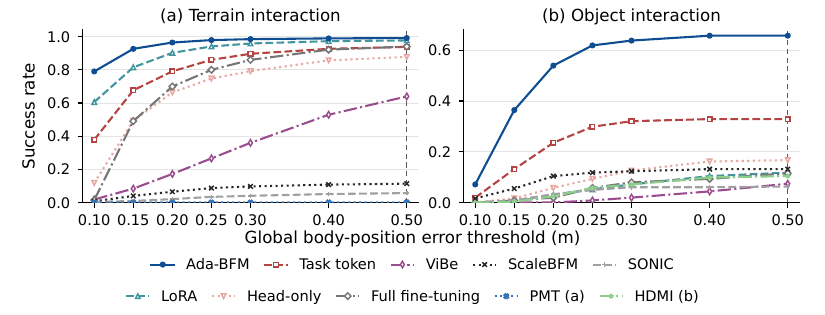}
\end{center}
\caption{Success rate as a function of the global body position error threshold.
Ada-BFM outperforms the baselines on both
the terrain and object interaction test sets.}
\label{fig:threshold-sweep}
\end{figure}

\section{Experiments}
\label{sec:experiments}

\subsection{Experimental Setup}
\label{sec:experimental-setup}

\paragraph{Environment setup.}
We conduct experiments with the Unitree G1 humanoid. We train Ada-BFM in
Isaac Lab on eight NVIDIA RTX 4090 GPUs, with 4096 environments per GPU.
We adapt a shared model for each interaction domain.
Training takes approximately $3.5$ hours for terrain interaction and $19$ hours
for object interaction under this configuration. 

\paragraph{Dataset.}
Terrain interaction uses data from OmniRetarget \citep{omniretarget2025} and
GRAIL \citep{grail2026}. Object interaction uses data retargeted from OMOMO
using OmniRetarget.
Appendix~\ref{app:datasets} describes the data used in each domain. We also evaluate on the BONES Test Set \citep{bones2026} to assess whether the
adapted policy preserves the original BFM's motion capabilities.

\paragraph{Baseline approaches.}
We compare against ScaleBFM and SONIC \citep{luo2026sonic}, full fine-tuning of
the pretrained BFM, training only the action head, LoRA \citep{hu2022lora},
Task Tokens \citep{vainshtein2025tasktokens} and ViBe
\citep{krishna2026vibe}. We also compare against PMT
\citep{wang2026perceptivebfm} for terrain interaction, and HDMI
\citep{hdmi2025} for object interaction. We use the public
checkpoints for ViBe, PMT, and HDMI. We train the remaining baselines
under the same configuration as Ada-BFM. Details of the metrics and success criteria are provided in
Appendix~\ref{app:eval-metrics}.

\needspace{13\baselineskip}
\subsection{Main Results}
\label{sec:main-results}
\noindent\begin{minipage}{\linewidth}
We evaluate Ada-BFM against three objectives
(Section~\ref{sec:introduction}): \textit{Effectiveness}, \textit{Efficiency},
and \textit{Retention}.

\begin{itemize}
\item \textbf{Effectiveness:} Can Ada-BFM support diverse tasks within each interaction domain using a single adapted model and outperform baseline approaches?
\item \textbf{Efficiency:} Can Ada-BFM leverage pretrained behavioral knowledge to learn scene-interaction capabilities faster than training from scratch?
\item \textbf{Retention:} Can Ada-BFM preserve the pretrained BFM's control capabilities while extending its ability to interact with scenes?
\end{itemize}
\end{minipage}

\paragraph{Effectiveness.}
To answer the first question, we adapt Ada-BFM separately for terrain and object
interaction and compare it with the baselines. Figure~\ref{fig:threshold-sweep}
plots success rate against the global body position error threshold on both
test sets. Tables~\ref{tab:track-terrain} and~\ref{tab:track-object} report
body and object tracking errors.

ScaleBFM and SONIC have low success rates on both interaction test sets
(Figure~\ref{fig:threshold-sweep}). They struggle to track motions in terrain
interaction (Table~\ref{tab:track-terrain}). For object interaction, they track
the reference body motion accurately but have large errors in object position
and rotation (Table~\ref{tab:track-object}). Accurate body motion tracking
alone is therefore insufficient for successful object interaction.

As shown in Figure~\ref{fig:threshold-sweep}, Ada-BFM achieves the highest success rate in both terrain and object interaction. It also achieves the lowest body tracking errors across all metrics in terrain interaction (Table~\ref{tab:track-terrain}). As the success criterion tightens in terrain interaction, baseline success rates decline sharply, while Ada-BFM maintains a high success rate. For object interaction, while Ada-BFM does not achieve the lowest errors on all body tracking metrics, it achieves the lowest errors on both object tracking metrics (Table~\ref{tab:track-object}) and outperforms these baseline methods. 

We further evaluate robustness to mismatches between the reference motion and the scene by applying random offsets to the terrain or object in $x$, $y$, $z$, and yaw while keeping the reference motion unchanged. For each perturbation level, we evaluate each reference motion with five randomly sampled sets of offsets and compute the success rate. Table~\ref{tab:scene-perturbation} shows that Ada-BFM can accommodate moderate mismatches between the reference motion and the scene, but success rates decline markedly as the offsets become large.

These results demonstrate that Ada-BFM supports diverse tasks within each interaction domain using a single adapted model and outperforms baseline approaches. Real-world experiments further show that Ada-BFM can interact with unseen terrains and objects, as illustrated in Figure~\ref{fig:teaser}.

\begin{table}[t]
\centering
{
\renewcommand{\arraystretch}{1.1}
\small
\setlength{\tabcolsep}{6pt}
\begin{tabular}{lcccc}
\hline
Method & G-MPKPE$\downarrow$ & G-MPKRE$\downarrow$ & L-MPKPE$\downarrow$ & L-MPKRE$\downarrow$ \\
\hline
ScaleBFM & 0.822 & 0.940 & 0.465 & 0.810 \\
SONIC & 0.968 & 0.632 & 0.385 & 0.608 \\
PMT & 1.712 & 1.517 & 0.639 & 1.259 \\
Full fine-tuning & 0.103 & 0.203 & 0.057 & 0.201 \\
Head-only & 0.145 & 0.287 & 0.081 & 0.283 \\
LoRA & 0.060 & 0.134 & 0.036 & 0.137 \\
ViBe & 0.258 & 0.185 & 0.088 & 0.177 \\
Task Tokens & 0.090 & 0.172 & 0.053 & 0.173 \\
\rowcolor{gray!12}
Ada-BFM & \textbf{0.042} & \textbf{0.081} & \textbf{0.023} & \textbf{0.084} \\
\hline
\end{tabular}
}
\captionof{table}{Body tracking errors on the terrain interaction test set. Ada-BFM outperforms all baselines across all metrics.}
\label{tab:track-terrain}

\vspace{1.2em}

{
\renewcommand{\arraystretch}{1.1}
\small
\setlength{\tabcolsep}{2.5pt}
\begin{tabular}{lcccccc}
\hline
 & \multicolumn{4}{c}{Body} & \multicolumn{2}{c}{Object} \\
\cline{2-7}
Method & G-MPKPE$\downarrow$ & G-MPKRE$\downarrow$ & L-MPKPE$\downarrow$ & L-MPKRE$\downarrow$ & G-MPKPE$\downarrow$ & G-MPKRE$\downarrow$ \\
\hline
ScaleBFM & 0.135 & \textbf{0.240} & 0.069 & 0.238 & 0.564 & 1.139 \\
SONIC & 0.318 & 0.255 & \textbf{0.054} & \textbf{0.230} & 0.590 & 1.075 \\
HDMI & 0.526 & 1.045 & 0.275 & 0.954 & 0.519 & 0.914 \\
Full fine-tuning & 0.428 & 0.864 & 0.217 & 0.838 & 0.544 & 1.115 \\
Head-only & 0.260 & 0.594 & 0.125 & 0.566 & 0.547 & 1.086 \\
LoRA & 0.403 & 0.855 & 0.220 & 0.823 & 0.571 & 1.155 \\
ViBe & 0.779 & 1.291 & 0.332 & 1.236 & 0.694 & 1.227 \\
Task Tokens & 0.121 & 0.256 & 0.064 & 0.251 & 0.448 & 1.006 \\
\rowcolor{gray!12}
Ada-BFM & \textbf{0.114} & 0.297 & 0.069 & 0.290 & \textbf{0.243} & \textbf{0.745} \\
\hline
\end{tabular}
}
\captionof{table}{Body and object tracking errors on the object interaction test set. While Ada-BFM does not achieve the lowest errors on all body tracking metrics, it achieves the lowest errors on both object tracking metrics.}
\label{tab:track-object}

\vspace{1.2em}

{
\renewcommand{\arraystretch}{1.1}
\small
\setlength{\tabcolsep}{6pt}
\begin{tabular}{lccccc}
\hline
Level & $x/y$ (m) & $z$ (m) & Yaw ($^\circ$) & Terrain SR$\uparrow$ & Object SR$\uparrow$ \\
\hline
0 & 0 & 0 & 0 & 0.9914 & 0.6575 \\
1 & $[-0.05, 0.05]$ & $[-0.025, 0]$ & $[-5, 5]$ & 0.9886 & 0.6038 \\
2 & $[-0.10, 0.10]$ & $[-0.050, 0]$ & $[-10, 10]$ & 0.9663 & 0.5485 \\
3 & $[-0.20, 0.20]$ & $[-0.100, 0]$ & $[-20, 20]$ & 0.8681 & 0.3436 \\
\hline
\end{tabular}
}
\captionof{table}{Robustness to scene pose perturbations. Ada-BFM test-set success rates (SR) under random translation and yaw offsets applied to terrains or objects, with reference motions unchanged. Each motion is evaluated five times per level, with offsets randomly sampled from their respective intervals for each evaluation. Level 0 uses no offsets.}
\label{tab:scene-perturbation}
\end{table}

\begin{figure}[t]
\begin{center}
\includegraphics[width=1.374in]{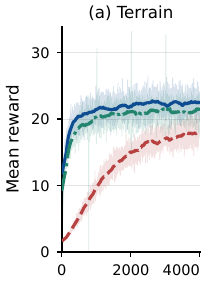}\hspace{0.03in}%
\includegraphics[width=1.202in]{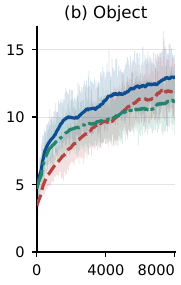}\hspace{0.12in}%
\includegraphics[width=1.404in]{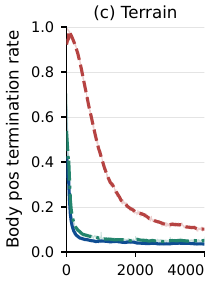}\hspace{0.03in}%
\includegraphics[width=1.232in]{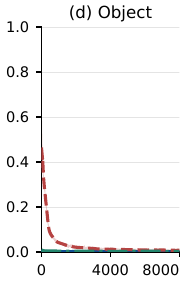}\\[3pt]
\includegraphics[width=\linewidth]{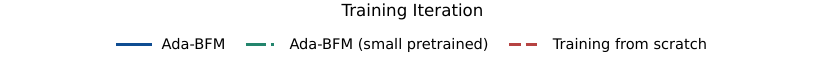}
\end{center}
\caption{Training comparison of Ada-BFM, adaptation from a small pretrained BFM and training from scratch. (a, b) Ada-BFM converges faster and achieves a higher
reward than training from scratch in both interaction domains. (c, d) Both pretrained
BFM variants have fewer terminations caused by body position error early in
training than the from-scratch baseline.}
\label{fig:training-curves}
\end{figure}

\begin{table}[t]
\captionsetup{singlelinecheck=false,justification=raggedright}
\begin{center}
\renewcommand{\arraystretch}{1.2}
\setlength{\tabcolsep}{8pt}
\begin{tabular}{lcc}
\hline
Method & Test Set SR (Terrain)$\uparrow$ & Test Set SR (Object)$\uparrow$ \\
\hline
\rowcolor{gray!12}
Ada-BFM & \textbf{0.9914} & \textbf{0.6575} \\
Ada-BFM (small pretrained) & 0.7819 & 0.4164 \\
Training from scratch & 0.0041 & 0.1151 \\
\hline
\end{tabular}
\end{center}
\caption{Test-set success rates of Ada-BFM, Ada-BFM on a small pretrained BFM, and training from scratch.}
\label{tab:narrower-ablation}
\end{table}


\paragraph{Efficiency.}

To answer the second question, we examine whether pretrained motion priors accelerate learning of humanoid scene interaction and whether stronger motion priors lead to faster learning and better task performance. We compare three methods for learning terrain and object
interaction tasks: Ada-BFM, Ada-BFM (small pretrained), and training from
scratch. For Ada-BFM (small pretrained), we select another ScaleBFM
checkpoint with smaller model capacity, pretrained on a smaller reference motion
dataset covering a narrower range of behaviors. For training from scratch, we
train a BFM using the same training configuration without loading pretrained
parameters.

As shown in Figure~\ref{fig:training-curves}(a, b), Ada-BFM converges faster
and achieves a higher reward than training from scratch.
These results show that pretrained motion priors accelerate learning and improve scene interaction policies.
The termination curves in Figure~\ref{fig:training-curves}(c, d) help explain
this advantage. Ada-BFM reuses pretrained motion control capabilities and
experiences fewer early terminations due to body position error. Training
from scratch requires learning motion tracking alongside scene interaction.
The pretrained prior therefore reduces this learning burden during adaptation.

As shown in Figure~\ref{fig:training-curves}(a, b) and Table~\ref{tab:narrower-ablation}, Ada-BFM converges faster and achieves higher rewards than Ada-BFM (small pretrained), while also attaining higher success rates on both the terrain and object interaction test sets. The results suggest that a higher-capacity BFM pretrained on a more diverse motion dataset provides a stronger foundation for learning humanoid scene interaction.

As shown in Figure~\ref{fig:training-curves}(a, b), Ada-BFM provides a larger improvement over training from scratch in terrain interaction than in object interaction, both in convergence speed and final reward. We attribute this difference in part to the training objectives. Terrain interaction focuses on accurately tracking reference motions, which aligns closely with the pretrained BFM's objective. Object interaction additionally requires interacting with objects, introducing demands beyond motion tracking that may require more exploration before convergence.

\begin{table}[t]
\begin{center}
\renewcommand{\arraystretch}{1.2}
\setlength{\tabcolsep}{18pt}
\begin{tabular}{lcc}
\hline
Method & BONES SR$\uparrow$ (Terrain) & BONES SR$\uparrow$ (Object) \\
\hline
Full fine-tuning & 0.734 \succdrop{0.238} & 0.408 \succdrop{0.564} \\
Head-only & 0.875 \succdrop{0.097} & 0.465 \succdrop{0.507} \\
LoRA & 0.849 \succdrop{0.123} & 0.351 \succdrop{0.621} \\
ViBe & 0.539 \succdrop{0.433} & 0.271 \succdrop{0.701} \\
Task Tokens & 0.936 \succdrop{0.036} & 0.929 \succdrop{0.043} \\
\rowcolor{gray!12}
Ada-BFM & 0.944 \succdrop{0.028} & 0.893 \succdrop{0.079} \\
\hline
\end{tabular}
\end{center}
\caption{Retention after adaptation on the BONES Test Set. Success uses a global body position threshold
of $0.5$~m. The Terrain and Object columns report results for the checkpoint
adapted to each domain. Frozen ScaleBFM achieves $0.972$ on the same clips.}
\label{tab:bones-retention}
\end{table}

\begin{figure}[t]
\begin{center}
\includegraphics[width=\linewidth]{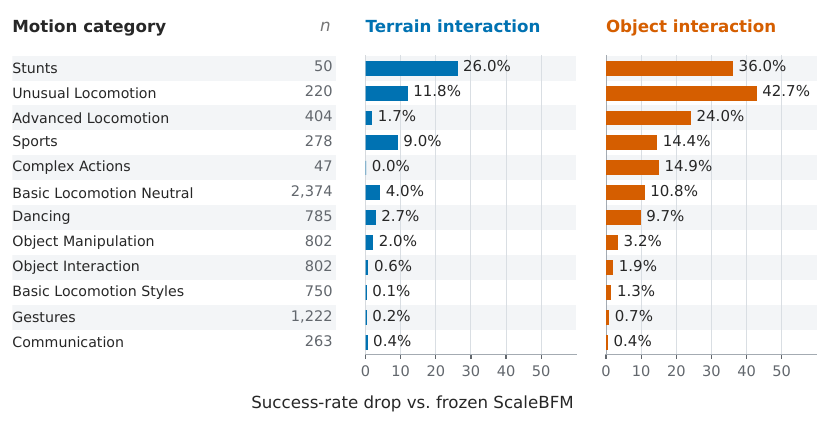}
\end{center}
\caption{Retention drop on the BONES Test Set by motion category. Bars give the
decrease in success rate (percentage points) of the terrain- and object-adapted
checkpoints relative to the frozen ScaleBFM, under the success criterion of
Table~\ref{tab:bones-retention}. $n$ is the number of clips.}
\label{fig:bones-categories}
\end{figure}

\paragraph{Retention.}
To answer the third question, we evaluate Ada-BFM on the BONES Test Set, with the
scene pathway enabled. As shown in Table~\ref{tab:bones-retention}, the
terrain- and object-adapted policies exhibit relatively small decreases in
success rate. These findings show that
Ada-BFM extends the pretrained BFM to humanoid scene interaction while largely
preserving its original motion-control capabilities. Retention is stronger
after terrain adaptation. As discussed in the efficiency analysis, this
pattern is consistent with the alignment of terrain interaction with the
pretrained motion-tracking objective, whereas object interaction introduces
additional demands beyond tracking.

Ada-BFM and Task Tokens preserve motion capabilities well through additional
conditioning pathways with frozen backbones. LoRA instead updates the
backbone's effective weights: it retains much of the pretrained capability
after terrain adaptation but suffers severe degradation after object
adaptation, alongside low success on
held-out object-interaction motions (Figure~\ref{fig:threshold-sweep}).
These results support separate conditioning pathways for balancing
interaction learning and capability retention.

Figure~\ref{fig:bones-categories} shows success rate decreases across
motion categories in the BONES Test Set, identifying the capabilities
most affected by adaptation.
Gestures, communication, object manipulation, object interaction, and basic
locomotion styles remain well preserved, whereas dancing shows moderate
degradation. The largest losses occur in Stunts and Unusual Locomotion,
which include flips, rolls, and crawling. Limited coverage of these
behaviors during adaptation and altered visual observations under unusual
body poses may contribute to their greater degradation.

\paragraph{Real-world deployment.}
We deploy the checkpoints adapted for terrain and object interaction on a
Unitree G1 equipped with a RealSense D435i depth camera.
We adapt a shared model for each interaction domain. All motions and scenes used in the real-world demos are unseen during training.

\begin{figure}[H]
\begin{center}
\includegraphics[width=\linewidth]{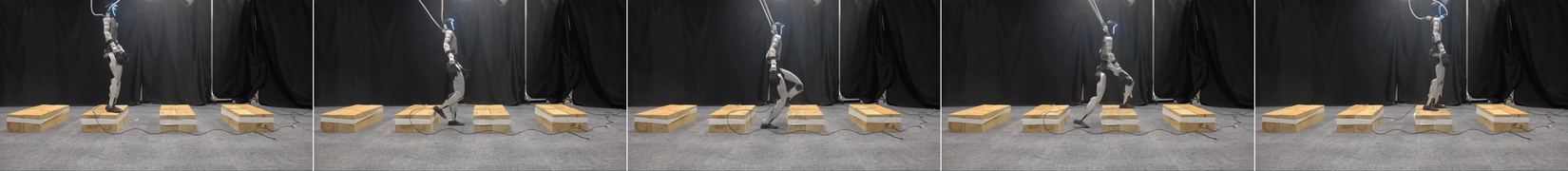}
\end{center}
\caption{Misstep and recovery during a curb traversal.}
\label{fig:recovery}
\end{figure}

The terrain and object checkpoints support diverse interaction behaviors on hardware (Figure~\ref{fig:teaser}), including lifting a box containing more than 3~kg of water (Figure~\ref{fig:teaser}(j)). The policy can also recover from a misstep: in a curb trial, it steps into empty space, regains its balance, and continues (Figure~\ref{fig:recovery}).

\begin{table}[H]
\captionsetup{singlelinecheck=false,justification=raggedright}
\begin{center}
\renewcommand{\arraystretch}{1.2}
\setlength{\tabcolsep}{8pt}
\begin{tabular}{lcc}
\hline
Method & Test Set SR (Terrain)$\uparrow$ & Test Set SR (Object)$\uparrow$ \\
\hline
\rowcolor{gray!12}
Ada-BFM & \textbf{0.9914} & \textbf{0.6575} \\
w/o FFN & 0.9314 & 0.3041 \\
w/o attention & 0.9882 & 0.4548 \\
Ada-BFM (zero-depth) & 0.9898 & 0.4685 \\
\hline
\end{tabular}
\end{center}
\caption{Adapter component and depth observation ablations. Ada-BFM achieves the highest test-set success rates in both interaction domains.}
\label{tab:adapter-component-ablation}
\end{table}

\subsection{Ablation Study}
\label{sec:ablation-studies}

We examine the contributions of the adapter components and depth observations
using success rates on the terrain and object interaction test sets.

\paragraph{Adapter components.}
Table~\ref{tab:adapter-component-ablation} reports test-set success rates after
removing either the adapter feed-forward network (FFN) or its attention branch.
Both ablations reduce success in both domains, but removing the FFN causes
a larger drop than removing attention. The degradation is more pronounced in
object interaction. These results demonstrate the contribution of
both components.

\paragraph{Depth observations.}
To assess the role of depth observations in both domains, we train and evaluate
a zero-depth variant with depth inputs fixed to zero throughout training and
deployment. As shown in Table~\ref{tab:adapter-component-ablation}, object
interaction success drops substantially, consistent with the
importance of depth cues for locating objects.  Appendix~\ref{app:depth-role} shows that depth enables smoother terrain interaction and more accurate tracking than the zero-depth baseline.

\section{Conclusion}
\label{sec:conclusion}

We presented Ada-BFM, a post-training framework for adapting pretrained
behavior foundation models to humanoid scene interaction. Ada-BFM supports diverse tasks with a shared model for each
interaction domain. Experiments demonstrate diverse terrain and object interaction behaviors, faster learning than training from scratch, and substantial retention of pretrained motion capabilities. These findings suggest that Ada-BFM is a
promising post-training framework for incorporating
external information and extending BFMs beyond their pretrained capabilities.

\section{Acknowledgments}
This work was supported by the Shanghai Municipal Special Program for Basic Research on General AI Foundation Models (Grant No. 2025SHZDZX025D08), in collaboration with Shanghai Artificial Intelligence Laboratory. The SJTU team is partially supported by National Natural Science Foundation of China (62322603).

\clearpage

%
%
%
\clearpage

\bibliography{iclr2027_conference}
\bibliographystyle{iclr2027_conference}

\clearpage
\appendix
\raggedbottom
\section{Experimental Details}
\label{app:experimental-details}
\subsection{Evaluation Metrics and Success Criteria}
\label{app:eval-metrics}

We report success rate (SR) and global and local tracking errors. At each
time step, the body tracking error is the mean Euclidean distance between
the tracked body links and their reference positions in the global frame.
Unless otherwise specified, a motion is successfully tracked if the maximum
of this error over the entire sequence does not exceed $0.5$\,m. This criterion
applies to flat-ground tracking and terrain interaction. For object interaction,
we additionally require the maximum Euclidean distance between the object
and its reference position in the global frame to be at most $0.4$\,m over
the sequence. Both the body and object position criteria must be satisfied.
SR is the fraction of evaluated motions that satisfy the applicable criteria.
For the success-rate curves that vary the body tracking threshold, the object
position threshold remains fixed at $0.4$\,m.

G-MPKPE and G-MPKRE measure the per-keypoint position and rotation errors of
selected links in the global frame, respectively. L-MPKPE and L-MPKRE measure
the corresponding root-relative errors after removing horizontal translation
offsets and heading differences. Position errors are measured in meters and
rotation errors in radians.

\subsection{Dataset Details}
\label{app:datasets}

\paragraph{Terrain interaction.}
We use the robot--terrain sequences from OmniRetarget and the curb, slope,
and stairs sequences from GRAIL.

\paragraph{Object interaction.}
We use data retargeted from OMOMO using OmniRetarget. Specifically, we use
only sequences involving the largebox object. Both the training and
test sets for object interaction are drawn exclusively from this subset.

For each interaction domain, we randomly split the clips into $80\%$ for
training and $20\%$ for testing.

\subsection{Depth Preprocessing}
\label{app:depth-preprocessing}

Following the preprocessing design of Hiking in the Wild~\citep{hiking2026},
both interaction domains use a buffer of four depth frames. At each episode reset,
we sample a temporal offset from $\{0,1,2,3\}$ frames. We read the corresponding
frame at each control step to model sensing latency. Each selected frame is
cropped by two pixels per side and resized bilinearly to $18\times32$.
Depth values are clipped to $[0,2.5]$~m and normalized to $[0,1]$ in both terrain
and object interaction. The buffer supplies a delayed observation rather than
a stack of depth frames to the encoder.

\subsection{Reward Settings}
\label{app:rewards}

Following ScaleBFM~\citep{zeng2026scaling}, we use a weighted sum of tracking
rewards, penalty terms, and a survival reward. Terrain interaction retains
these terms with unchanged weights and tracking scales. Object interaction
adds object position, object orientation, and contact agreement rewards.
Table~\ref{tab:base_rewards} summarizes all components. The reward at each
control step is
\begin{equation}
 r_t^k = \Delta t\sum_{i\in\mathcal{I}_k} w_i r_{i,t},
\end{equation}
where $\mathcal{I}_k$ contains the shared terms for terrain interaction and
additionally the object interaction terms for object interaction.
$\Delta t$ is the control timestep.

\begin{table}[!t]
\centering
\small
\setlength{\tabcolsep}{4pt}
\renewcommand{\arraystretch}{1.5}
\begin{tabular*}{\linewidth}{@{\extracolsep{\fill}}lcrr@{}}
\toprule
Term & Expression & Weight & Notes \\
\midrule
\multicolumn{4}{c}{Tracking Rewards (Both Domains)} \\
\midrule
Pelvis height
 & $\exp\!\left(-\frac{(h_t-h_t^{\mathrm{ref}})^2}{\sigma_h^2}\right)$
 & $0.5$ & $\sigma_h=0.3$ \\
Global body position
 & $\exp\!\left(-\frac{\sum_{b\in\mathcal{B}}\|p_{t,b}-p_{t,b}^{\mathrm{ref}}\|_2^2}{|\mathcal{B}|\sigma_p^2}\right)$
 & $1.0$ & $\sigma_p=0.3$ \\
Global body orientation
 & $\exp\!\left(-\frac{\sum_{b\in\mathcal{B}}d_R(q_{t,b},q_{t,b}^{\mathrm{ref}})^2}{|\mathcal{B}|\sigma_q^2}\right)$
 & $1.0$ & $\sigma_q=0.4$ \\
Global body linear velocity
 & $\exp\!\left(-\frac{\sum_{b\in\mathcal{B}}\|v_{t,b}-v_{t,b}^{\mathrm{ref}}\|_2^2}{|\mathcal{B}|\sigma_v^2}\right)$
 & $1.0$ & $\sigma_v=1.0$ \\
Global body angular velocity
 & $\exp\!\left(-\frac{\sum_{b\in\mathcal{B}}\|\omega_{t,b}-\omega_{t,b}^{\mathrm{ref}}\|_2^2}{|\mathcal{B}|\sigma_\omega^2}\right)$
 & $1.0$ & $\sigma_\omega=3.14$ \\
\midrule
\multicolumn{4}{c}{Penalty Terms (Both Domains)} \\
\midrule
Action rate & $\|a_t-a_{t-1}\|_2^2$ & $-0.1$ & --- \\
Joint position limits
 & $\sum_j[(\ell_j-\theta_{j,t})_+ +(\theta_{j,t}-u_j)_+]$
 & $-10.0$ & Soft limits \\
\midrule
\multicolumn{4}{c}{Survival Reward (Both Domains)} \\
\midrule
Survival & $\mathbb{1}[\mathrm{is\_alive}]$ & $1.0$ & --- \\
\midrule
\multicolumn{4}{c}{Object Interaction Rewards (Object Domain Only)} \\
\midrule
Global object position
 & $\exp\!\left(-\frac{\|p_t^{\mathrm{obj}}-p_t^{\mathrm{obj,ref}}\|_2^2}{\sigma_{\mathrm{op}}^2}\right)$
 & $2.0$ & $\sigma_{\mathrm{op}}=0.15$ \\
Global object orientation
 & $\exp\!\left(-\frac{d_R(q_t^{\mathrm{obj}},q_t^{\mathrm{obj,ref}})^2}{\sigma_{\mathrm{oq}}^2}\right)$
 & $1.0$ & $\sigma_{\mathrm{oq}}=0.6$ \\
Contact agreement & $r_{\mathrm{contact},t}$ & $1.0$ & Force $>1$\,N \\
\bottomrule
\end{tabular*}
\caption{Reward components and weights. Superscript $\mathrm{ref}$ denotes
reference motion quantities; quantities without it describe the simulated
state. Tracking scales use meters, radians, meters per second, or radians
per second as appropriate. All weights are applied before multiplication
by the control timestep $\Delta t$.}
\label{tab:base_rewards}
\end{table}

Here $h$ denotes pelvis height; $p$, $q$, $v$, and $\omega$ denote global
position, orientation, linear velocity, and angular velocity, respectively.
$d_R$ is the angle of the relative rotation between two orientations.
The tracked set $\mathcal{B}$ contains the pelvis, torso, and bilateral
hip-roll, knee, ankle-roll, shoulder-roll, elbow, and wrist-yaw links.
Squared body tracking errors are averaged over these 14 links before applying the
exponential. $a_t$ is the policy action, $\theta_{j,t}$ is joint position,
and $[\ell_j,u_j]$ gives its soft limits, with $(x)_+=\max(x,0)$.
The survival indicator is one when no non-timeout termination occurs.

\paragraph{Contact agreement.}
We evaluate contact at the left and right palm pads and ankle-roll links,
denoted by $\mathcal{C}$. For each link, $c_{\ell,t}=1$ when the maximum force
magnitude over its robot--object contact pairs exceeds $1$\,N. Given reference
contact labels $c_{\ell,t}^{\mathrm{ref}}$, the reward is
\begin{equation}
 r_{\mathrm{contact},t} =
 \begin{cases}
 \frac{1}{|\mathcal{C}|}\sum_{\ell\in\mathcal{C}}
 \mathbb{1}[c_{\ell,t}=c_{\ell,t}^{\mathrm{ref}}],
 & \text{if reference labels are available},\\[0.4em]
 \mathbb{1}[\exists\ell\in\mathcal{C}:c_{\ell,t}=1],
 & \text{otherwise}.
 \end{cases}
 \label{eq:contact_reward}
\end{equation}
This term rewards both contact and absence of contact when they match the
reference. Without contact labels, it rewards contact at any of the four
links. The contact reward is additive and does not gate the object pose
rewards. Object states and contact measurements are used for reward
computation during training; they are not required as actor inputs at deployment.

\subsection{Domain Randomization}
\label{app:domain-randomization}

During post-training, we randomize physical parameters and depth observations
to expose the policy to variation in contact dynamics, robot parameters, and
sensing conditions. These variations are intended to reduce dependence on a
single nominal simulation configuration when transferring to the real robot.
Tables~\ref{tab:domain-randomization} and~\ref{tab:observation-randomization}
report the randomization settings for terrain and object
interaction. Following the presentation in ScaleBFM~\citep{zeng2026scaling},
we group the settings by physical parameters and perturbation type.
$\mathcal{U}[a,b]$ denotes a uniform distribution. Offsets are additive unless
specified as multiplicative scales. Settings apply to both interaction domains
unless marked otherwise.

\paragraph{Physical parameters and disturbances.}
Both domains randomize robot friction, default joint positions, and torso
center of mass to represent variation in contact conditions and robot
calibration. Object interaction additionally varies object dimensions, mass,
and friction, exposing the policy to changes in geometry and load during
manipulation. The two domains use different robot friction ranges, as listed
in Table~\ref{tab:domain-randomization}. Physical parameters are sampled at
initialization, whereas linear and angular root velocity increments are
applied periodically during rollouts to perturb the robot's balance.

\begin{table}[H]
\centering
\small
\setlength{\tabcolsep}{4pt}
\renewcommand{\arraystretch}{1.2}
\begin{tabular}{@{}p{0.30\linewidth}p{\dimexpr0.70\linewidth-2\tabcolsep\relax}@{}}
\toprule
    Domain Randomization & Distribution / Setting \\
\midrule
\multicolumn{2}{l}{\textit{Robot and object material settings}} \\
Material sampling & 64 sampled buckets \\
Friction constraint & Object interaction only: sampled dynamic friction is capped by static friction \\
\midrule
\multicolumn{2}{l}{\textit{Robot physical parameters}} \\
Static friction & Terrain: $\mu_s\sim\mathcal{U}[0.3,1.6]$; object: $\mathcal{U}[0.6,1.2]$ \\
Dynamic friction & Terrain: $\mu_d\sim\mathcal{U}[0.3,1.2]$; object: $\mathcal{U}[0.4,1.0]$ \\
Default joint positions & $q_0\leftarrow q_0+\mathcal{U}[-0.01,0.01]$ rad \\
Torso COM offset & $\Delta x\sim\mathcal{U}[-0.025,0.025]$ m; $\Delta y,\Delta z\sim\mathcal{U}[-0.05,0.05]$ m \\
\midrule
\multicolumn{2}{l}{\textit{Object physical parameters (object interaction only)}} \\
Object shape scale & Dimensions scaled along the three axes: $s_x,s_y,s_z\sim\mathcal{U}[0.9,1.1]$ \\
Object mass scale & $m\leftarrow m_0 s_m$, $s_m\sim\mathcal{U}[0.8,1.2]$ \\
Object static friction & $\mu_s^{\mathrm{obj}}\sim\mathcal{U}[0.5,1.3]$ \\
Object dynamic friction & $\mu_d^{\mathrm{obj}}\sim\mathcal{U}[0.3,1.1]$ \\
\midrule
\multicolumn{2}{l}{\textit{Root velocity perturbations}} \\
Linear velocity increments & $\Delta v_x,\Delta v_y\sim\mathcal{U}[-0.5,0.5]$ m/s; $\Delta v_z\sim\mathcal{U}[-0.2,0.2]$ m/s \\
Angular velocity increments & $\Delta\omega_x,\Delta\omega_y\sim\mathcal{U}[-0.52,0.52]$ rad/s; $\Delta\omega_z\sim\mathcal{U}[-0.78,0.78]$ rad/s \\
Perturbation interval & $T\sim\mathcal{U}[5,10]$ s \\
\bottomrule
\end{tabular}
\caption{Physical randomization and root velocity perturbations.
Physical parameters are sampled once at initialization. Root velocity
increments are applied at intervals $T$.}
\label{tab:domain-randomization}
\end{table}

\paragraph{Camera and depth observations.}
Camera pose perturbations represent uncertainty in sensor mounting, while
depth delay, measurement noise, and zero-valued patches represent latency,
range errors, and missing measurements, respectively.
Camera extrinsics are sampled at initialization; the depth delay is sampled
at each episode reset and remains fixed within that episode.
The policy receives a single delayed depth frame at each control step,
using the preprocessing described in Appendix~\ref{app:depth-preprocessing}.
Table~\ref{tab:observation-randomization} lists the corresponding distributions.

\begin{table}[H]
\centering
\small
\setlength{\tabcolsep}{4pt}
\renewcommand{\arraystretch}{1.2}
\begin{tabular}{@{}p{0.30\linewidth}p{\dimexpr0.70\linewidth-2\tabcolsep\relax}@{}}
\toprule
Domain Randomization & Sampling Distribution \\
\midrule
\multicolumn{2}{l}{\textit{Camera extrinsics}} \\
Camera translation & $\Delta x,\Delta y,\Delta z\sim\mathcal{U}[-0.01,0.01]$ m \\
Camera orientation & Roll/yaw: $\mathcal{U}[-2^{\circ},2^{\circ}]$; pitch: $\mathcal{U}[-10^{\circ},10^{\circ}]$ \\
\midrule
\multicolumn{2}{l}{\textit{Depth observations}} \\
Depth delay & $d\sim\operatorname{Uniform}\{0,1,2,3\}$ frames \\
Depth measurement noise & $\epsilon\sim\mathcal{N}(0,0.02^2)$ m, applied to depths in $[0.12,1.2]$ m \\
Depth artifacts & Zero-valued patches; seed probability $10^{-4}$ per valid pixel; height/width sampled from $\mathcal{N}(2,0.5^2)$ pixels and clipped to image bounds \\
\bottomrule
\end{tabular}
\caption{Camera and depth randomization in the current default
configuration. Camera extrinsics are randomized once at initialization.
Depth delay is sampled at episode reset and held fixed within the episode.}
\label{tab:observation-randomization}
\end{table}

\subsection{Control Modes}
\label{app:control-modes}

We inherit the eight control modes of ScaleBFM~\citep{zeng2026scaling}.
Each mode specifies the body links whose reference motion is provided as
motion goals, ranging from sparse target specifications to whole-body control.
Table~\ref{tab:control-modes} lists the activated links using the zero-based
mode indices in our implementation. In the main experiments, all motions used for training and evaluation are conditioned
on mode~7 (Whole-Body Mode), with all 14 target links activated. This setting
applies to post-training in both interaction domains, evaluation on their
held-out test sets, and retention evaluation on the BONES Test Set. The control mode remains fixed in these experiments.

\begin{table}[H]
\centering
\small
\setlength{\tabcolsep}{4pt}
\renewcommand{\arraystretch}{1.35}
\begin{tabular*}{\linewidth}{@{\extracolsep{\fill}}clcp{0.43\linewidth}@{}}
\toprule
Mode & Control Mode & \# Links & Activated Links \\
\midrule
0 & Root & 1 & Pelvis \\
1 & Bimanual & 2 & Left hand, right hand \\
2 & Root-and-Hand & 3 & Pelvis, left hand, right hand \\
3 & End-Effector & 4 & Left hand, right hand, left foot, right foot \\
4 & Root-and-End-Effector & 5 & Pelvis, left hand, right hand, left foot, right foot \\
5 & Upper-Body & 6 & Left/right shoulder, elbow, and hand \\
6 & Root-and-Upper-Body & 7 & Pelvis, left/right shoulder, elbow, and hand \\
7 & Whole-Body & 14 & Pelvis, torso, left/right shoulder, elbow, hand, hip, knee, and foot \\
\bottomrule
\end{tabular*}
\caption{Control modes inherited from ScaleBFM. Mode indices run from 0 to 7;
\# Links denotes the number of activated target links. The main training and evaluation experiments use mode~7, including the interaction test sets
and the BONES retention test.}
\label{tab:control-modes}
\end{table}

In the robot implementation, hand, foot, shoulder, and hip targets correspond
to the wrist-yaw, ankle-roll, shoulder-roll, and hip-roll links, respectively.

%

\section{Role of Depth Perception in Terrain Interaction}
\label{app:depth-role}

We compare depth-conditioned and zero-depth policies using tracking-error
distributions and averages of per-motion peak tracking errors to examine
benefits beyond the small
terrain success-rate difference in Table~\ref{tab:adapter-component-ablation}.
The zero-depth policy receives zero depth inputs during training and deployment.

\paragraph{Peak tracking errors.}
For each test motion, we compute the maximum frame-wise tracking error over
the entire sequence, where the frame-wise error is the mean global position
error over the tracked body links. We then report percentiles of these
per-motion maxima in Table~\ref{tab:depth-tail-errors}. Without terrain offsets,
the median peak error is $0.067$\,m with depth and $0.072$\,m with zero depth, a difference
of only $5$\,mm. In contrast, depth reduces p99 from $0.728$\,m to $0.426$\,m,
a reduction of approximately $30$\,cm. Thus, the benefit is concentrated in
the tail of the error distribution rather than in its typical values.

Visual inspection suggests that the motor capabilities inherited from the
pretrained BFM allow the zero-depth policy to recover from some errors caused
by mismatches between reference motions and terrain, such as inaccurate foot
placement leading to missed steps and stumbles. The depth-conditioned policy
largely avoids these errors in the inspected rollouts. This observation suggests why similar success rates can coexist with
differences in tracking stability.

To further examine whether depth helps handle mismatches between reference
motions and terrain, we evaluate both policies with terrain offsets of
$\pm0.05$\,m along $x$ and $y$, $[-0.025,0]$\,m along $z$, and $\pm5^{\circ}$
in yaw. Under these perturbations, depth reduces p90 from $1.547$\,m to
$0.589$\,m (Table~\ref{tab:depth-tail-errors}). These results suggest that depth helps compensate for reference--terrain
mismatches, although large errors remain at p99 for both policies.

\begin{table}[htbp]
  \centering
  \small
  \begin{tabular*}{\linewidth}{@{\extracolsep{\fill}}lrrr@{}}
    \toprule
    Percentile & Depth (m) & Zero-depth (m) & Relative increase \\
    \midrule
    \multicolumn{4}{l}{\textit{No terrain offsets}} \\
    p50 & 0.067 & 0.072 & $+7.5\%$ \\
    p75 & 0.092 & 0.099 & $+7.6\%$ \\
    p90 & 0.134 & 0.145 & $+8.2\%$ \\
    p99 & 0.426 & 0.728 & $+70.9\%$ \\
    \midrule
    \multicolumn{4}{l}{\textit{With terrain position and orientation offsets}} \\
    p50 & 0.195 & 0.210 & $+7.7\%$ \\
    p75 & 0.262 & 0.302 & $+15.3\%$ \\
    p90 & 0.589 & 1.547 & $+162.6\%$ \\
    p99 & 2.805 & 3.018 & $+7.6\%$ \\
    \bottomrule
  \end{tabular*}
  \caption{Percentiles of per-motion maximum frame-wise tracking error in
  terrain interaction. Lower is better.}
  \label{tab:depth-tail-errors}
\end{table}

\paragraph{Average peak tracking errors.}
Table~\ref{tab:depth-mean-errors} reports the average per-motion peak error
for each of the four tracking metrics. For each motion and each metric, we
first compute the frame-wise error and take its maximum over the entire
sequence; we then average these maxima across test motions. Thus, these
values summarize per-motion worst-case errors, rather than time-averaged
tracking errors.
G-MPKPE/G-MPKRE measure global per-keypoint position/rotation errors;
L-MPKPE/L-MPKRE measure their root-relative counterparts after removing
horizontal translation offsets and heading differences. Position and rotation
errors use meters and radians, respectively. Relative to the
depth-conditioned policy, the zero-depth policy incurs $5.9\%$--$9.9\%$
higher errors across these metrics. Depth thus improves tracking quality even when the aggregate success rate
changes little.

\begin{table}[htbp]
  \centering
  \small
  \begin{tabular*}{\linewidth}{@{\extracolsep{\fill}}lrrr@{}}
    \toprule
    Metric & Depth & Zero-depth & Relative increase \\
    \midrule
    G-MPKPE (m)   & 0.0952 & 0.1046 & $+9.9\%$ \\
    L-MPKPE (m)   & 0.0403 & 0.0432 & $+7.2\%$ \\
    G-MPKRE (rad) & 0.1698 & 0.1799 & $+5.9\%$ \\
    L-MPKRE (rad) & 0.1743 & 0.1886 & $+8.2\%$ \\
    \bottomrule
  \end{tabular*}
  \caption{Average per-motion peak tracking errors in terrain interaction
  with and without depth observations. For each metric, we take the maximum
  frame-wise error over each motion and then average across test motions.
  Lower is better.}
  \label{tab:depth-mean-errors}
\end{table}

\end{document}